\documentclass[runningheads]{llncs}
\usepackage[T1]{fontenc}
\usepackage{graphicx}
\usepackage{amsfonts}                   
\usepackage{amsmath}                    
\usepackage{amssymb}                    

\usepackage[hang]{subfigure}            

\graphicspath{{images/}}
\usepackage{subfiles} 
\usepackage{hyperref}

\usepackage[disable]{todonotes}
\usepackage{blindtext}

\newcommand*{\newlinecommand}[2]{%
  \newcommand*{#1}{%
    \begingroup%
    \escapechar=`\\%
    \catcode\endlinechar=\active%
    \csname\string#1\endcsname%
  }%
  \begingroup%
  \escapechar=`\\%
  \lccode`\~=\endlinechar%
  \lowercase{%
    \expandafter\endgroup
    \expandafter\def\csname\string#1\endcsname##1~%
  }{\endgroup#2\space}%
}

\newlinecommand{\x}{\todo[inline, size=\small, color=blue!20]{$\bullet$ #1}}
\newlinecommand{\xx}{\todo[inline, size=\small, color=green!20]{\noindent\hspace*{3mm} $\circ$ #1}}
\newlinecommand{\xxx}{\todo[inline, size=\small, color=white!20]{\noindent\hspace*{5mm} -- #1}}

\usepackage{color}

\newcommand{\bte}[1]{%
}
\newcounter{blindtextcounter}

\begin{document}
\title{Bring the Noise: Introducing Noise Robustness to Pretrained Automatic Speech Recognition}
\titlerunning{Introducing Noise Robustness to Pretrained Automatic Speech Recognition}
\author{Patrick Eickhoff\inst{1} \and
Matthias M\"oller\inst{2} \and
Theresa Pekarek Rosin\inst{1} \and
Johannes Twiefel\inst{1,3} \and
Stefan Wermter\inst{1}}
\authorrunning{P. Eickhoff et al.}
%
\institute{Knowledge Technology, Department of Informatics, University of Hamburg,
Vogt-Koelln-Str. 30, 22527 Hamburg, Germany\\
\email{\{patrick.eickhoff, theresa.pekarek-rosin, johannes.twiefel, stefan.wermter\}@uni-hamburg.de}\\
\url{www.knowledge-technology.info}\and
Centre for Applied Autonomous Sensor Systems (AASS), \"Orebro University, Sweden\\
\email{matthias.moeller@oru.se}\\
\and
exXxa GmbH, Vogt-Koelln-Str. 30, 22527 Hamburg, Germany\\
\url{www.exXxa.ai}}
\maketitle              
\begin{abstract}
\todo{maybe change title to something more catchy?}
In recent research, in the domain of speech processing, large End-to-End (E2E) systems for Automatic Speech Recognition (ASR) have reported state-of-the-art performance on various benchmarks. These systems intrinsically learn how to handle and remove noise conditions from speech. Previous research has shown, that it is possible to extract the denoising capabilities of these models into a preprocessor network, which can be used as a frontend for downstream ASR models. However, the proposed methods were limited to specific fully convolutional architectures. In this work, we propose a novel method to extract the denoising capabilities, that can be applied to any encoder-decoder architecture. We propose the Cleancoder preprocessor architecture that extracts hidden activations from the Conformer ASR model and feeds them to a decoder to predict denoised spectrograms. We train our preprocessor on the Noisy Speech Database (NSD) to reconstruct denoised spectrograms from noisy inputs. Then, we evaluate our model as a frontend to a pretrained Conformer ASR model as well as a frontend to train smaller Conformer ASR models from scratch. We show that the Cleancoder is able to filter noise from speech and that it improves the total Word Error Rate (WER) of the downstream model in noisy conditions for both applications.
\todo{also keeping the performance for low-noise conditions}

\todo{Automatic Speech Recognition, ask Stefan here}
\end{abstract}
\section{Introduction}
End-to-End (E2E) systems have been the state-of-the-art approach to ASR for a few years now \cite{baevskiWav2vecFrameworkSelfSupervised2020,w2vbert,radfordRobustSpeechRecognition}. An E2E system usually takes audio as input, processes it into an internal representation, and produces a transcript of the speech. The big advantage of these systems is that all components are trained together, so they can learn a joint representation. However, the disadvantage is that they often require deep models with a large number of parameters to perform well. For example, the recent Whisper-large model~\cite{radfordRobustSpeechRecognition} contains about 1550~M parameters. Training a model like this from scratch is computationally expensive and usually not possible for research institutions. However, most of these models can be successfully adapted to smaller domains through the use of transfer learning, which indicates the quality of speech representations learned~\cite{hsuHuBERTSelfSupervisedSpeech2021,radfordRobustSpeechRecognition}. Additionally, E2E systems usually do not have preprocessing applied to their input and the model itself has to learn how to separate speech from noise~\cite{radfordRobustSpeechRecognition}. Usually, the earlier layers of recent ASR architectures are required to separate noise from speech implicitly. 
When a new ASR architecture is developed, the earlier noise-handling layers need to be trained again. 
This raises the question if it is possible to separate the processing capabilities of such a large and powerful pretrained ASR model and reuse them for another model.

Our work takes inspiration from the recent findings of Möller et al. \cite{moeller2021}. They were able to utilize a pre-trained Jasper \cite{liJasperEndtoEndConvolutional2019} ASR model to create a preprocessor, which increases the noise robustness of pretrained models and improves the performance of smaller ASR models trained from scratch. However, their approach has two disadvantages: 1) their approach is only applicable to a specific architecture of Jasper, and 2) Jasper is no longer state-of-the-art for English ASR. Instead, attention-based models derived from the Transformer \cite{Vaswani2017AttentionIA} architecture have outperformed convolutional and recurrent ASR approaches \cite{gulatiConformerConvolutionaugmentedTransformer2020,hsuHuBERTSelfSupervisedSpeech2021,radfordRobustSpeechRecognition,chenWavLMLargeScaleSelfSupervised2022}. Therefore, we propose a new extraction method (Parallel Weighted Sum), which is potentially applicable to any encoder-decoder ASR architecture. We apply this method to a Conformer \cite{gulatiConformerConvolutionaugmentedTransformer2020} model, a state-of-the-art attention-based architecture, to create our own preprocessor called Cleancoder.
Our model can function either as an independent frontend for pre-trained ASR models or can be used in combination with architectures trained from scratch to improve their noise robustness. In our experiments, we measure the performance (Word Error Rate; WER) on different noise levels on the Noisy Speech Dataset (NSD)~\cite{valentini-botinhaoNoisySpeechDatabase2017} to show that our methods improve the performance under noisy conditions and the performance does not decrease under clean conditions (LibriSpeech~\cite{panayotovLibrispeechASRCorpus2015}) when using our preprocessor. 



\section{Related Work}
To improve the noise robustness of a speech recognition model, training processes usually include adding both artificial and realistic noise to the training data. This leads to large-scale ASR models showcasing a certain noise robustness without any further preprocessing steps. However, since smaller models might not be able to perform the same internal denoising steps, it is important to examine how the capabilities of larger models can be exploited by smaller models.

There are different approaches to creating external preprocessors, that denoise speech for further speech recognition. Many of these focus on filtering noise from speech using statistical methods in combination with deep learning methods~\cite{Caroselli2022,fang2023partially,Heymann2016beamformer}. 
A recent example of such methods is the Cleanformer model. Cleanformer~\cite{Caroselli2022} is a multichannel frontend architecture for speech enhancement based on the Conformer~\cite{gulatiConformerConvolutionaugmentedTransformer2020} architecture. 
The model combines raw noisy speech and enhanced input features, produced by a SpeechCleaner~\cite{Huang2019} noise cancellation algorithm, to create an Ideal Ratio Mask (IRM)~\cite{irm}. This mask estimates the ratio of speech in the noisy signal in the spectral space.
These ratios are then applied to the input signal to filter out noise. The model works independently of the combined ASR model and is able to reduce WERs across multiple SNR values by approximately 50~\%.

Instead of applying a filtering method to the nosiy signal, our approach reconstructs clean spectrograms completely from latent representations. Our work is based on the findings of Möller et al.~\cite{moeller2021}, who created a frontend architecture for noise filtering based on the Jasper~\cite{liJasperEndtoEndConvolutional2019} architecture. Based on the findings of Li et al.~\cite{liWhatDoesNetwork2020} and their probing methods to extract and predict spectrograms from hidden representations, Möller et al. applied this method to gauge the denoising capabilities of a pre-trained Jasper model.
The underlying assumption is that areas of speech that the model perceives as noise are filtered out very early by the system and are not represented in the model's latent space. Thereby, those filtering capabilities could be leveraged for other ASR models and increase their noise robustness.


Möller et al. demonstrate how the reconstructed representations of speech support other already pretrained ASR systems in noisy conditions. Additionally, they observe that those features support other ASR systems as input while training and that models with those representations generally perform better on noisy and clean data. 
However, their approach relies strongly on the architecture of Jasper \cite{liJasperEndtoEndConvolutional2019} and its residual connections. They retrain the batch normalization layers in the model and are therefore limited to one specific architecture, which is not state-of-the-art anymore. Our work introduces a way that could potentially  reconstruct speech from any ASR system while still retaining denoising capabilities. 


\section{Methodology}
Our own method of constructing a denoising preprocessor from a pretrained ASR model is inspired by the work of Möller et al. \cite{moeller2021}. However, we propose an architecture that is able to extract latent representation from potentially any encoder-decoder ASR architectures and is not limited to the Jasper architecture. Our Cleancoder model extracts the latent representations of an ASR model's encoder and reconstructs denoised spectrograms.
\begin{figure}[htb]
    \centering
    \includegraphics[width=\textwidth]{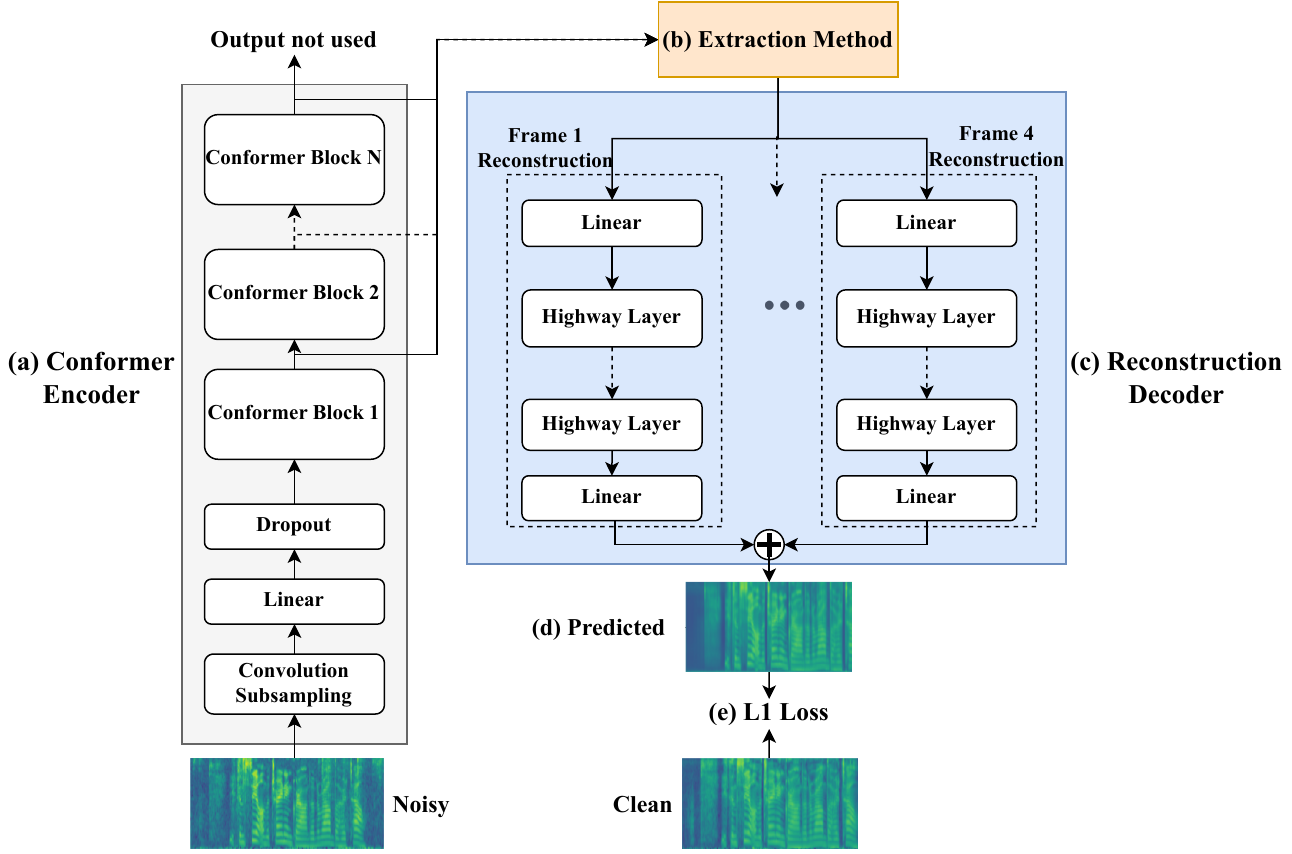}
    \caption{This figure shows the architecture of our Cleancoder. On the left, we display the original Conformer encoder architecture (a). Then, the output of every Conformer block is fed into our extraction method (b). This method computes a weighted sum of the latent representation and feeds this vector to our reconstruction decoder (c). This decoder contains four different Highway Networks tasked with reconstructing one fourth of the final output frame. 
    The four outputs are appended along the temporal axis to reconstruct a complete spectrogram of a frame (d). Then we compute the L1 loss (e) between the reconstructed spectrogram and the clean ground truth.}
    \label{fig:ownarchitecture}
\end{figure}

The architecture follows an encoder-decoder structure, shown in \autoref{fig:ownarchitecture}. We choose pretrained Conformer models as a baseline to extract our preprocessor from it. These models are larger than the Jasper~\cite{liJasperEndtoEndConvolutional2019} used by Möller et al. \cite{moeller2021} but still can be trained from scratch in reasonable time. There are multiple pretrained Conformer models with CTC available from NVIDIA NeMo \footnote{\url{https://catalog.ngc.nvidia.com/models}}. 
 Jasper has been outperformed by many more recent ASR models, which are using attention-based architectures \cite{w2vbert,chung2021w2v,gulatiConformerConvolutionaugmentedTransformer2020}.
The Jasper model used by Möller et al. only reported a WER of 2.84\%(test-clean)/7.84\%(test-other)~\cite{liJasperEndtoEndConvolutional2019} on the LibriSpeech \cite{panayotovLibrispeechASRCorpus2015} testset, using an external language model. The large Conformer we base our model on reported 1.9\%(test-clean)/3.9\%(test-other)~\cite{gulatiConformerConvolutionaugmentedTransformer2020}. Thus, we assume that our Cleancoder will be able to yield better WER improvements for downstream models. 

\begin{figure}
    \centering
    \includegraphics[width=0.8\textwidth]{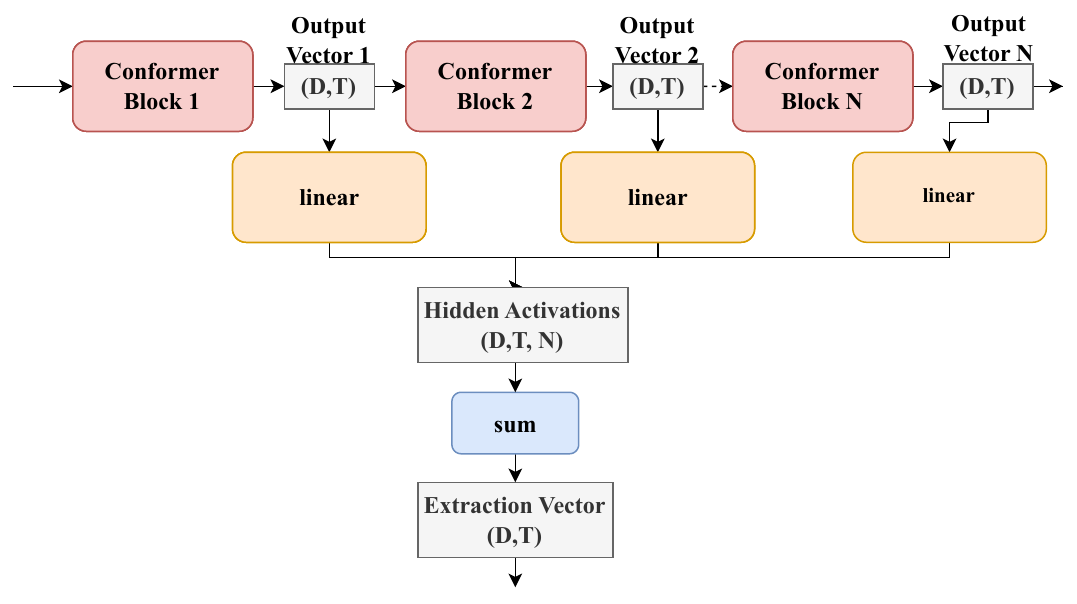}
    \caption{This figure shows our Parallel Weighted Sum method. Each latent vector of the Conformer encoder has size (D,T) and is fed through a separate fully connected layer after which we sum all the projected vectors into one output vector of size (D,T).}
    \label{fig:wsum}
\end{figure}

We reuse the encoder of the ASR Conformer model (see \autoref{fig:ownarchitecture}, a). We disregard the decoder layers as we're only interested in the latent representations. To extract these hidden activations, we feed the output of each Conformer block into our extraction method.
We create an approach that is applicable to potentially any encoder-decoder ASR architecture while remaining simple. 
We propose a Parallel Weighted Sum extraction method (see \autoref{fig:wsum}), which extends the regular weighted-sum. 
However, instead of choosing one weight vector to reduce all the hidden activations into one, our method feeds each layer through separate parallel projection layers and computes the sum across these layers. 
This way, we do not only weigh the contribution of the different blocks to the denoised output but also weigh the information contained in each feature vector. 
We took inspiration from the work of Yang et al. \cite{yangSUPERBSpeechProcessing2021}, who compared different Self-Supervised Learned (SSL) representations. 

For our decoder network, we choose to follow the example of Möller et al. \cite{moeller2021} and use four-layer Highway Networks~\cite{srivastavaHighwayNetworks2015}. They have shown, that these networks are able to reconstruct spectrograms sufficiently for ASR from hidden representations. Since the Conformer preprocessing block reduces the temporal dimension by a factor of four, we train four different Highway Networks. 
 The four outputs are appended along the temporal axis. Given the input $x$ consists of $t$ frames, the Conformer will reduce the temporal dimension by four yielding $t/4$ frames. We denote the latent representation constructed by our Parallel Weighted Sum as $s_i$ for frame $i$ and our four Highway Networks as $N_1,N_2,\dots , N_4$. Since our decoder is almost identical to the one of Möller et al. \cite{moeller2021}, we obtain a similar equation for the output $y$ of our model: 
	
\begin{equation*}
    y = (N_1(s_0),N_2(s_0),N_3(s_0),N_4(s_0),\dots ,N_1(s_{t/4}),N_2(s_{t/4}),N_3(s_{t/4}),N_4(s_{t/4}))
\end{equation*}

\section{Experimental Results}

\subsection{Datasets}
\subsubsection{Noisy Speech Database}
    The Noisy Speech Database (NSD) \cite{valentini-botinhaoNoisySpeechDatabase2017} was designed to test and train speech enhancement algorithms. It contains pairs of noisy and clean speech, sampled at 48kHz, and is divided into a training and test set. For our experiments, we downsampled our input to 16kHz. Each sample in the datasets provides noisy and clean audio, a transcript, information about the speaker, signal-to-noise ratio (SNR), and noise type. There are two sets of the NSD with 28 \cite{valentini-botinhaoSpeechEnhancementNoisy2018} and 56 speakers \cite{nsd56} taken from the Voice Bank Corpus \cite{veauxVoiceBankCorpus2013} \todo{both datasets differ in the spoken accent, check this}. The noisy samples were created by adding recorded noise from the DEMAND database \cite{thiemannDEMANDCollectionMultichannel2013} as well as generated babble and speech-shaped noise. These noise types were applied at different SNRs. We combine the 28 and 56 speaker sets to expose our model to a larger variety of speakers and noise conditions. Thus, we will ensure better generalization. 
    We use the NSD to train our denoising preprocessor and to evaluate the performance of downstream ASR models on noisy data.

\subsubsection{LibriSpeech}
	LibriSpeech \cite{panayotovLibrispeechASRCorpus2015} is a corpus of approximately 1000 hours of clean English speech, sampled at 16kHz. 
 LibriSpeech is an established dataset for evaluating ASR models \cite{gulatiConformerConvolutionaugmentedTransformer2020,liJasperEndtoEndConvolutional2019}. We use LibriSpeech in our experiments 
 to train small ASR models from scratch.

\subsection{Training the Cleancoder}
	\label{subsec:training}
    To evaluate if the Cleancoder architecture filters noise from speech we train two preprocessor models (medium, large) on the NSD train set. This way, we can estimate the required size of the best preprocessor.
    Our preprocessors are trained to reconstruct spectrograms of the same form as the encoder's input. 
    These are log-Mel spectrograms with 80 features, a window size of 0.025, and a windows stride of 0.01. We convert each clean and noisy audio signal of each sample in the NSD trainset into log-Mel spectrograms.
    While training, our models are fed the noisy spectrograms and predict denoised spectrograms. 

   \begin{figure}[htb]
		\centering
		\includegraphics[scale=0.44]{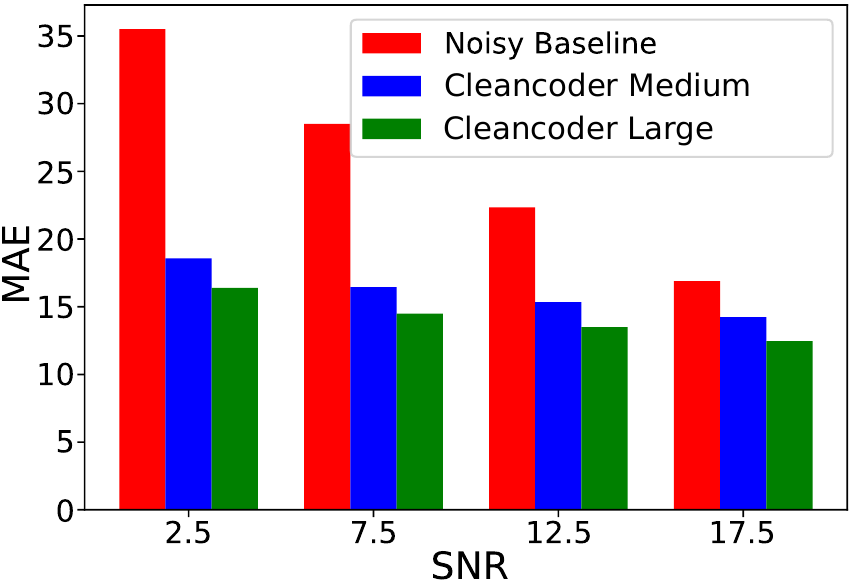}
		\caption{This figure presents the MAE computed between the noisy and clean and respective denoised and clean spectrograms of the NSD testset. We observe that the denoised spectrograms of both preprocessors show a lower MAE than the noisy baseline across all noise conditions. The large Cleancoder shows the lowest MAE.}
		\label{fig:frontendmae}
	\end{figure}
    
    Then, we can compute the L1 loss between the clean and denoised spectrograms.
    We train two different models with the medium and large sized Conformer CTC models. The medium Conformer consists of \~ 30.7M parameters, while the large one consists of \~ 118.8M parameters \cite{gulatiConformerConvolutionaugmentedTransformer2020}.
	For each encoder model, we train our preprocessor for 100 epochs on a batch size of 64 with L1 Loss. The learning rate is set to a magnitude of $1e^{-3}$, where the precise values are taken from a hyperparameter search, which we conduct prior to the actual training. The Adam optimizer is configured with $\beta_1=0.9$, $\beta_2=0.98$ and a weight decay of $1e^{-4}$. The learning rates are set to the optimal values from our hyperparameter search. We choose to omit the learning rate scheduler, since the initial learning rate is already very small. Our decoder is configured as four four-layer Highway Networks.  	

 After training the two models, we inspect the differences between the noisy, clean, and denoised spectrograms. We measure the deviation of them by computing the mean absolute error (MAE) between the clean and noisy as well as clean and denoised spectrograms. Our results on the MAE are shown in \autoref{fig:frontendmae}. For both preprocessors, we observe that they reduce the MAE compared to just the noisy input. The lower the SNR the larger the improvement, indicating that the Cleancoder models filter noise from speech.

\subsection{Frontend for Pretrained Models}
Next, we test how our Cleancoder affects the performance of existing pretrained ASR models. Therefore, we use our preprocessors as frontends to first denoise the input signal and generate spectrograms. These are fed into a pretrained downstream ASR model which predicts transcriptions. Finally, we measure the WER between the ground truth texts, the transcripts recognized from the unprocessed noisy spectrograms (noisy baseline), and the transcripts recognized from the preprocessed noisy spectrograms (our preprocessor). 

	\begin{figure}[htb]
		\centering
		\subfigure[Conformer CTC ASR model]{\includegraphics[width=0.495\textwidth]{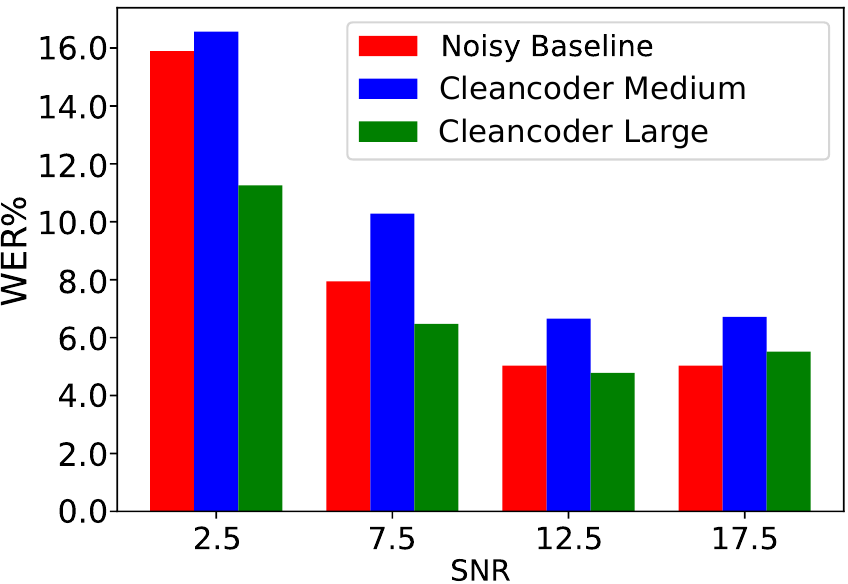}}
        \subfigure[Conformer Transducer ASR model]{\includegraphics[width=0.495\textwidth]{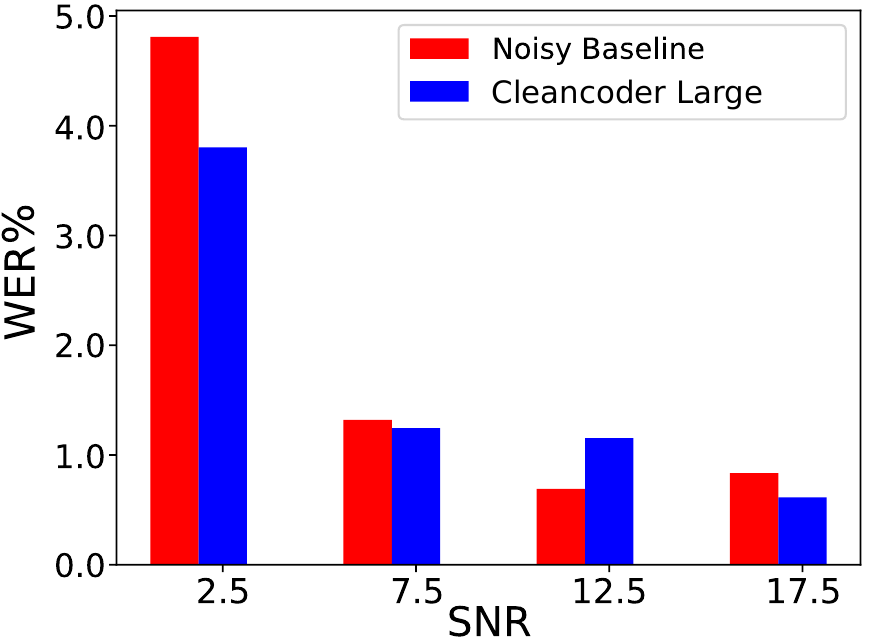}}
		\caption{Figure (a) presents the WER on the NSD testset for evaluating our frontends using a Conformer CTC downstream model. Figure (b) presents the WER using a Conformer Transducer downstream model. The Transducer was only evaluated with the large Cleancoder, as the medium Cleancoder had already proven to be unable to improve ASR performance. Both plots show the WER grouped by SNR. Each bar denotes either the noisy spectrograms or respective denoised spectrograms. The WER improves the most on low SNR samples and slightly degrades WER on high SNR samples.}
		\label{fig:frontendwernsd}
	\end{figure}

For our experiments, we choose a medium-sized Conformer with CTC and a large Conformer Transducer as downstream ASR models, which are both publicly available through NVIDIA's model collection. We chose two different ASR models to ensure a degree of invariance to the downstream architecture. This experiment verifies if it is possible to combine our frontend with other downstream architectures without the risk of degrading the performance.

 Our results are shown in \autoref{fig:frontendwernsd}. We can see, that overall, while the WER increases with the medium preprocessor compared to the baseline, it decreases for almost all SNR configurations with the large Cleancoder. The large Cleancoder performs better on samples with low SNR, only for samples with the highest SNR of 17.5 the performance is slightly worse than the noisy baseline. We observe that the performance of the baseline Conformer Transducer got worse from SNR 12.5 to 17.5. When analyzing the errors we found minor anomalies in the predictions, however, since the WER is already very low we accredit this observation to general variance.	


	 We further discuss the correlation of our MAE and WER results. 
  Möller et al. \cite{moeller2021} suggested, that the MAE and WER do not necessarily correlate. In fact, we found little research on the impact of the MAE between noisy and clean speech on the resulting WER. While we observe significant improvements of the MAE using our medium and large preprocessors, the WER shows significantly lower performance on the medium preprocessor. Only the large version yields positive results. There seems to be no strong correlation between the MAE and WER. We assume that a different loss function to train the preprocessor would be more appropriate, and we will examine this in our future work.


\subsection{Training an ASR Model from Scratch}

We evaluate how the Cleancoder impacts the training of a smaller downstream ASR model from scratch. The architecture of choice for the ASR model was a small Conformer using CTC without a language model.
We train three different small Conformer models. All three are trained on LibriSpeech's training splits. The baseline model uses no frontend, while the others are trained on the outputs of our medium and large preprocessor, respectively. All three are trained for 100 epochs with CTC loss using a batch size of 128 and an Adam optimizer with $\beta_1$ as 0.9 and $\beta_2$ as 0.98. We apply a NoamAnnealing learning rate scheduler with 10,000 warmup steps, an initial learning rate of 2.0, and a minimal learning rate of $1.0e^{-06}$.

\begin{figure}[htb]
		\centering
		\subfigure[LibriSpeech]{\includegraphics[width=0.50\textwidth]{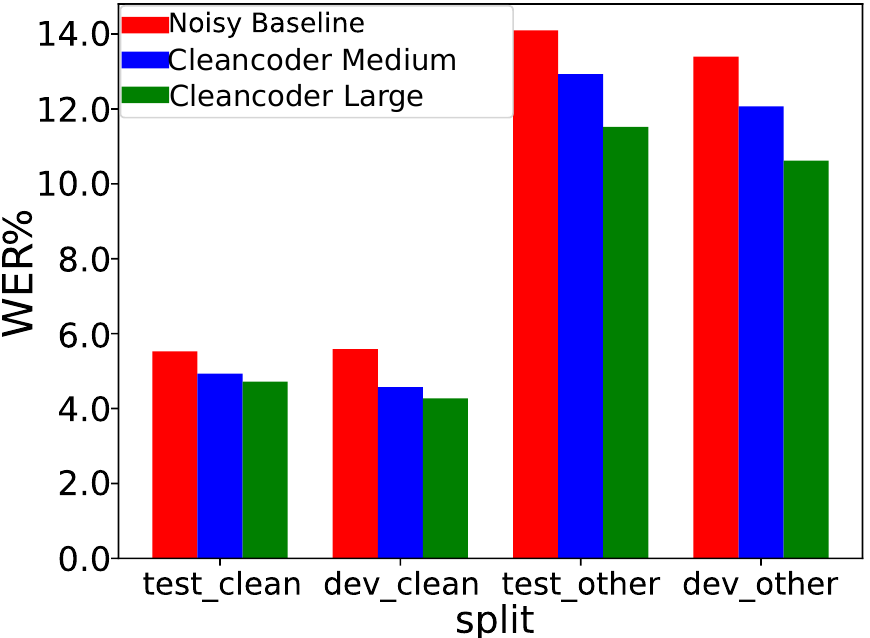}}
		\subfigure[NSD]{\includegraphics[width=0.485\textwidth]{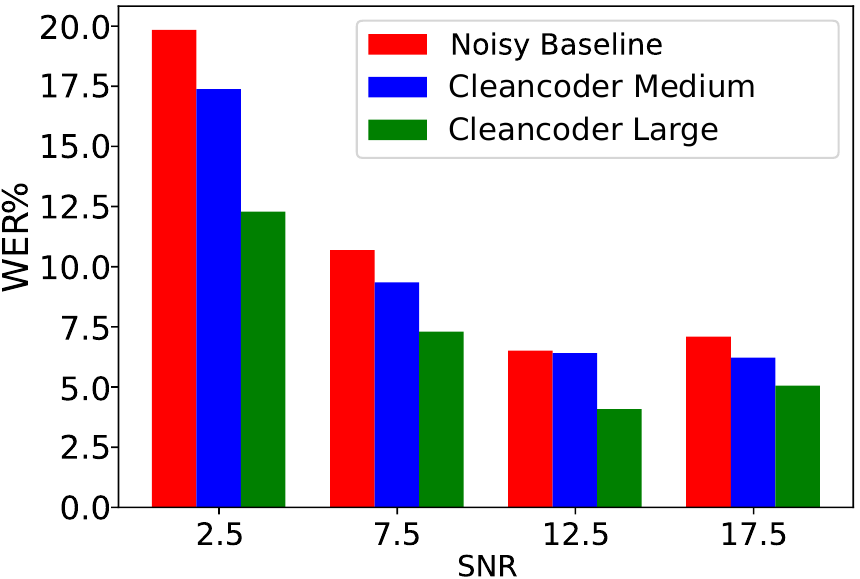}}
		\caption{Figure (a) shows the WER (\%) of our three ASR models trained from scratch for the dev- and test-splits of LibriSpeech. Figure (b) shows the WER (\%) for the NSD testset grouped by SNR. Each model used the same small Conformer architecture and was trained on either the raw input or the output of our medium and large preprocessors. We observe that the ASR model using the large preprocessor shows the lowest WER in both figures.}
		\label{fig:scratchwernsd}
\end{figure}

In the first plot of \autoref{fig:scratchwernsd}, we report the WER computed on the test-clean, test-other, dev-clean, and dev-other splits of LibriSpeech. We observe that both models using our preprocessors outperform the baseline ASR model. The large Cleancoder shows the lowest mean WER on all splits. 

Furthermore, we show the WER of our ASR models on the NSD testset grouped by SNR in the second plot of \autoref{fig:scratchwernsd}. 
 We observe that both models using our preprocessors outperform the baseline model. The large preprocessor shows the best performance with an almost 4\% improvement overall. We also observe that using the Cleancoders yields the biggest improvements on samples with a low SNR. This shows the models are more robust to noise.  

\begin{figure}[htb]
    \centering
    \subfigure[Validation CTC Loss]{\includegraphics[width=0.495\textwidth]{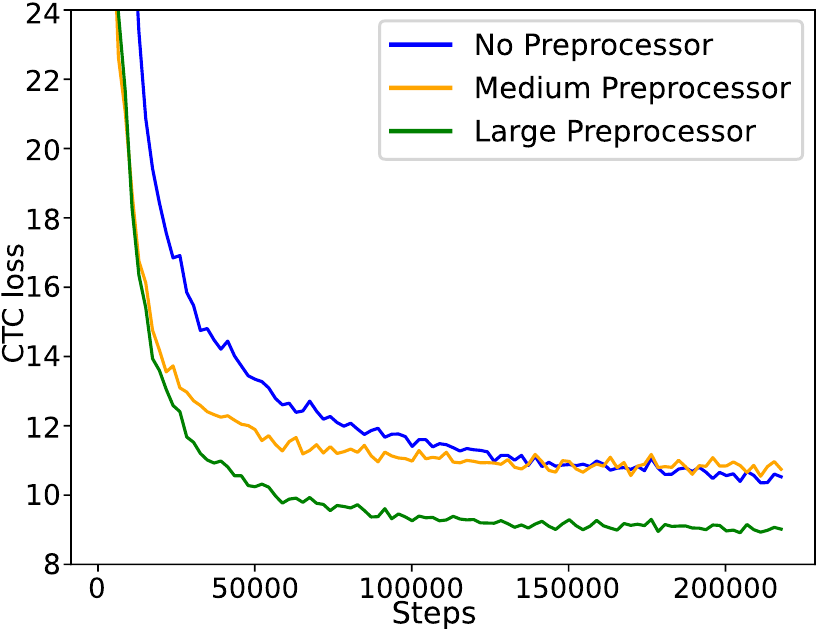}}
    \subfigure[Validation WER]{\includegraphics[width=0.495\textwidth]{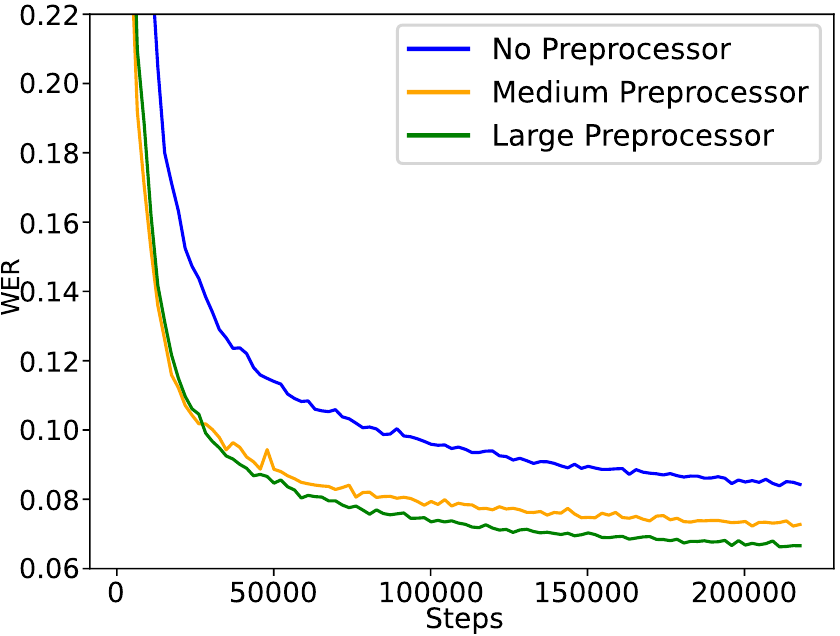}}\\
    \caption{
    Plot (a) shows the CTC loss over the number of training steps for the validation dataset. Plot (b) shows the WER, computed on the validation dataset, over the training steps. Each plot shows three curves for the baseline ASR model (blue), the medium preprocessor (orange), and the large preprocessor (green). Both preprocessors converge lower than the baseline ASR model, except for the CTC validation loss of the medium preprocessor.
    }
    \label{fig:scratchtrainval}
\end{figure}

 Finally, we investigate the impact of using our Cleancoders over the course of training. The evaluation CTC loss and evaluation WER are shown in \autoref{fig:scratchtrainval}. We observe that both models using our frontend converge faster and reach a lower loss and WER than the baseline model. The loss curves and WER curves follow an almost identical course. We observe that the validation loss and WER are both lowest for the large Cleancoder. As previously discussed, this supports the assumption that the large Cleancoder generalizes better to different noise conditions.


\section{Conclusion}

We created preprocessors from pretrained Conformer \cite{gulatiConformerConvolutionaugmentedTransformer2020} ASR models by extracting the hidden activations and training a decoder to predict denoised spectrograms. In our experiments, we showed that our Cleancoder improves the performance (WER) under noisy conditions (SNR: 2.5 and 7.5) for two different downstream ASR models. Under clean audio conditions (SNR: 12.5 and 17.5) the performance stayed mostly stable (with one outlier). The results indicate that our preprocessor is capable of improving the performance of downstream ASR models under noisy conditions without the necessity of performing any training for the downstream ASR models.
In the second experiment, we trained the downstream ASR model from scratch by first feeding the audio training data through the Cleancoder and then using the generated spectrograms as training data for our downstream ASR model. The performance substantially improved under both noisy and clean audio conditions. Also, we measured the training and validation loss while performing the training. These results show that the training time of an ASR model can be reduced due to an improved convergence, and the performance can be increased when using our preprocessor as input to a downstream ASR model.


In future work, we plan to research different loss functions aside from MAE to train the Cleancoder for denoising. One example could be the Ideal Ratio Mask (IRM)~\cite{Narayanan2013}, which has recently been successfully utilized for a denoising frontend~\cite{Caroselli2022}. A loss function with better correlation to the downstream WER, could further improve our Cleancoder's performance as an ASR frontend. 
Furthermore, we will evaluate our preprocessor using additional downstream ASR architectures. Especially the combination of our preprocessor and the recent Whisper \cite{radfordRobustSpeechRecognition} model would be worth investigating. Finally, applying our approach to create denoising preprocessors from other architectures will confirm if our method works with any encoder-decoder ASR architecture.

\subsubsection{Acknowledgements} The authors gratefully acknowledge support from the German BMWK (SIDIMO), the DFG (CML, LeCAREbot) and the European Commission (TRAIL, TERAIS). 

%
%
%
\bibliographystyle{splncs04}
\bibliography{bring_the_noise}

\begin{thebibliography}{10}
\providecommand{\url}[1]{\texttt{#1}}
\providecommand{\urlprefix}{URL }
\providecommand{\doi}[1]{https://doi.org/#1}

\bibitem{baevskiWav2vecFrameworkSelfSupervised2020}
Baevski, A., Zhou, Y., Mohamed, A., Auli, M.: wav2vec 2.0: A framework for
  self-supervised learning of speech representations. Advances in Neural
  Information Processing Systems  \textbf{33},  12449--12460 (2020)

\bibitem{Caroselli2022}
Caroselli, J., Naranayan, A., O'Malley, T.: {Cleanformer: A Microphone Array
  Configuration-Invariant, Streaming, Multichannel Neural Enhancement Frontend
  for ASR}. ArXiv  \textbf{abs/2204.11933} (2022)

\bibitem{chenWavLMLargeScaleSelfSupervised2022}
Chen, S., Wang, C., Chen, Z., Wu, Y., Liu, S., Chen, Z., Li, J., Kanda, N.,
  Yoshioka, T., Xiao, X., et~al.: {WavLM}: {L}arge-scale {S}elf-supervised
  {P}re-training for {F}ull {S}tack {S}peech {P}rocessing. IEEE Journal of
  Selected Topics in Signal Processing  \textbf{16}(6),  1505--1518 (2022)

\bibitem{w2vbert}
Chung, Y.A., Zhang, Y., Han, W., Chiu, C.C., Qin, J., Pang, R., Wu, Y.:
  {w2v-BERT: Combining Contrastive Learning and Masked Language Modeling for
  Self-Supervised Speech Pre-Training}. In: IEEE Automatic Speech Recognition
  and Understanding Workshop (ASRU). pp. 244--250 (2021)

\bibitem{chung2021w2v}
Chung, Y.A., Zhang, Y., Han, W., Chiu, C.C., Qin, J., Pang, R., Wu, Y.:
  W2v-bert: Combining contrastive learning and masked language modeling for
  self-supervised speech pre-training. In: 2021 IEEE Automatic Speech
  Recognition and Understanding Workshop (ASRU). pp. 244--250. IEEE (2021)

\bibitem{fang2023partially}
Fang, H., Wittmer, N., Twiefel, J., Wermter, S., Gerkmann, T.: Partially
  adaptive multichannel joint reduction of ego-noise and environmental noise.
  arXiv preprint arXiv:2303.15042  (2023)

\bibitem{gulatiConformerConvolutionaugmentedTransformer2020}
Gulati, A., Qin, J., Chiu, C.C., Parmar, N., Zhang, Y., Yu, J., Han, W., Wang,
  S., Zhang, Z., Wu, Y., Pang, R.: {Conformer: Convolution-augmented
  Transformer for Speech Recognition}. In: Proc. Interspeech. pp. 5036--5040
  (Oct 2020)

\bibitem{Heymann2016beamformer}
Heymann, J., Drude, L., Haeb-Umbach, R.: Neural network based spectral mask
  estimation for acoustic beamforming. In: IEEE International Conference on
  Acoustics, Speech and Signal Processing (ICASSP). pp. 196--200 (2016)

\bibitem{hsuHuBERTSelfSupervisedSpeech2021}
Hsu, W.N., Bolte, B., Tsai, Y.H.H., Lakhotia, K., Salakhutdinov, R., Mohamed,
  A.: {Hubert: Self-supervised Speech Representation Learning by Masked
  Prediction of Hidden Units}. IEEE/ACM Transactions on Audio, Speech, and
  Language Processing  \textbf{29},  3451--3460 (2021)

\bibitem{Huang2019}
Huang, Y.A., Shabestary, T.Z., Gruenstein, A.: Hotword cleaner: Dual-microphone
  adaptive noise cancellation with deferred filter coefficients for robust
  keyword spotting. In: Proceedings of IEEE International Conference on
  Acoustics, Speech and Signal Processing (ICASSP). pp. 6346--6350 (2019)

\bibitem{liWhatDoesNetwork2020}
Li, C., Yuan, P., Lee, H.: What {D}oes a {N}etwork {L}ayer {H}ear? {A}nalyzing
  {H}idden {R}epresentations of {E}nd-to-{E}nd {ASR} through {S}peech
  {S}ynthesis. In: Proceedings of IEEE International Conference on Acoustics,
  Speech and Signal Processing (ICASSP). pp. 6434--6438. IEEE Press (May 2020)

\bibitem{liJasperEndtoEndConvolutional2019}
Li, J., Lavrukhin, V., Ginsburg, B., Leary, R., Kuchaiev, O., Cohen, J.M.,
  Nguyen, H., Gadde, R.T.: {Jasper: An End-to-End Convolutional Neural Acoustic
  Model}. In: Proc. Interspeech. pp. 71--75 (Sep 2019)

\bibitem{moeller2021}
M{\"o}ller, M., Twiefel, J., Weber, C., Wermter, S.: Controlling the noise
  robustness of end-to-end automatic speech recognition systems. In:
  Proceedings of the International Joint Conference on Neural Networks (IJCNN)
  (Jul 2021)

\bibitem{irm}
Narayanan, A., Wang, D.: Ideal ratio mask estimation using deep neural networks
  for robust speech recognition. In: IEEE International Conference on
  Acoustics, Speech and Signal Processing. pp. 7092--7096 (2013)

\bibitem{Narayanan2013}
Narayanan, A., Wang, D.: {Ideal Ratio Mask Estimation Using Deep Neural
  Networks for Robust Speech Recognition}. In: Proceedings of IEEE
  International Conference on Acoustics, Speech and Signal Processing (ICASSP).
  pp. 7092--7096. IEEE Press (2013)

\bibitem{panayotovLibrispeechASRCorpus2015}
Panayotov, V., Chen, G., Povey, D., Khudanpur, S.: {Librispeech: An ASR Corpus
  Based on Public Domain Audio Books}. In: Proceedings of IEEE International
  Conference on Acoustics, Speech and Signal Processing (ICASSP). pp.
  5206--5210. IEEE Press (Apr 2015)

\bibitem{radfordRobustSpeechRecognition}
Radford, A., Kim, J.W., Xu, T., Brockman, G., McLeavey, C., Sutskever, I.:
  {Robust Speech Recognition via Large-Scale Weak Supervision}. arXiv preprint
  arXiv:2212.04356  (2022)

\bibitem{srivastavaHighwayNetworks2015}
Srivastava, R.K., Greff, K., Schmidhuber, J.: Highway networks. arXiv preprint
  arXiv:1505.00387  (2015)

\bibitem{thiemannDEMANDCollectionMultichannel2013}
Thiemann, J., Ito, N., Vincent, E.: {{DEMAND}: A Collection of Multi-Channel
  Recordings of Acoustic Noise in Diverse Environments}  (2013)

\bibitem{valentini-botinhaoNoisySpeechDatabase2017}
{Valentini-Botinhao}, C.: {Noisy Speech Database for Training Speech
  Enhancement Algorithms and {{TTS}} Models}  (2017)

\bibitem{nsd56}
Valentini-Botinhao, C., Wang, X., Takaki, S., Yamagishi, J.: {Investigating
  RNN-based speech enhancement methods for noise-robust Text-to-Speech.} In:
  Speech Synthesis Workshop (SSW). pp. 146--152 (Sep 2016)

\bibitem{valentini-botinhaoSpeechEnhancementNoisy2018}
Valentini-Botinhao, C., Yamagishi, J.: {Speech Enhancement of Noisy and
  Reverberant Speech for Text-to-Speech}. IEEE/ACM Transactions on Audio,
  Speech, and Language Processing  \textbf{26}(8),  1420--1433 (2018)

\bibitem{Vaswani2017AttentionIA}
Vaswani, A., Shazeer, N., Parmar, N., Uszkoreit, J., Jones, L., Gomez, A.N.,
  Kaiser, L.u., Polosukhin, I.: {Attention is All you Need}. In: Guyon, I.,
  Luxburg, U.V., Bengio, S., Wallach, H., Fergus, R., Vishwanathan, S.,
  Garnett, R. (eds.) Advances in Neural Information Processing Systems.
  vol.~30. Curran Associates, Inc. (Dec 2017)

\bibitem{veauxVoiceBankCorpus2013}
Veaux, C., Yamagishi, J., King, S.: {The Voice Bank Corpus: Design, Collection
  and Data Analysis of a Large Regional Accent Speech Database}. In: Oriental
  COCOSDA held jointly with 2013 Conference on Asian Spoken Language Research
  and Evaluation (O-COCOSDA/CASLRE), International Conference. Institute of
  Electrical and Electronics Engineers (IEEE), United States (Nov 2013)

\bibitem{yangSUPERBSpeechProcessing2021}
Yang, S.W., Chi, P.H., Chuang, Y.S., Lai, C.I.J., Lakhotia, K., Lin, Y.Y., Liu,
  A.T., Shi, J., Chang, X., Lin, G.T., Huang, T.H., Tseng, W.C., tik Lee, K.,
  Liu, D.R., Huang, Z., Dong, S., Li, S.W., Watanabe, S., Mohamed, A., yi~Lee,
  H.: {SUPERB: Speech Processing Universal PERformance Benchmark}. In: Proc.
  Interspeech. pp. 1194--1198 (2021)

\end{thebibliography}

\end{document}